%% file: ott_114.tex
\documentclass[accepted]{uai2023} 

\usepackage[american]{babel}

\usepackage{natbib} 
    \renewcommand{\bibsection}{\subsubsection*{References}}
\usepackage{mathtools} 
\usepackage{booktabs} 
\usepackage{tikz} 
\usepackage{amssymb}
\usepackage{layouts}
\usepackage[toc,page]{appendix}

\usetikzlibrary{arrows}
\usetikzlibrary{fadings, positioning}
\usetikzlibrary{external}
\usetikzlibrary{positioning}
\usetikzlibrary{calc}

\newenvironment{talign*}
 {\csname align*\endcsname}
 {\endalign}
\newenvironment{talign}
{\align}
{\endalign}

\hypersetup{colorlinks=true, citecolor=black, linkcolor=black} 
\newcommand{\update}[1]{\textcolor{black}{#1}}

\title{Bayesian Numerical Integration with Neural Networks}

\author[1, 2]{Katharina~Ott}
\author[1]{Michael~Tiemann}
\author[2, 3]{Philipp~Hennig}
\author[4, 5]{Fran\c{c}ois-Xavier~Briol}
\affil[1]{%
Bosch Center for Artificial Intelligence,
Renningen, Germany
}
\affil[2]{%
University of Tübingen, Tübingen, Germany
}
\affil[3]{%
MPI for Intelligent Systems,
Tübingen, Germany
}
\affil[4]{%
Department of Statistical Science, University College
London, London, United Kingdom
}
\affil[5]{%
The Alan Turing Institute, London, United Kingdom
}
  
\begin{document}
\maketitle

\begin{abstract}
  Bayesian probabilistic numerical methods for numerical integration offer significant advantages over their non-Bayesian counterparts: they can encode prior information about the integrand, and can quantify uncertainty over estimates of an integral. However, the most popular algorithm in this class, Bayesian quadrature, is based on Gaussian process models and is therefore associated with a high computational cost. To improve scalability, we propose an alternative approach based on Bayesian neural networks which we call \emph{Bayesian Stein networks}.
  The key ingredients are a neural network architecture based on Stein operators, and an approximation of the Bayesian posterior based on the Laplace approximation. We show that this leads to orders of magnitude speed-ups on the popular Genz functions benchmark, and on challenging problems arising in the Bayesian analysis of dynamical systems, and the prediction of energy production for a large-scale wind farm.
\end{abstract}

\section{INTRODUCTION}

Integration is a core task in probabilistic machine learning. It is required to perform operations such as marginalizing out random variables, or computing normalization constants, predictive distributions, and posterior expectations. 
In this paper, we consider the computation of the integral of some function $f:\mathcal{X} \rightarrow \mathbb{R}$, where $\mathcal{X} \subseteq \mathbb{R}^d$, against some distribution $\Pi$ with (Lebesgue) density $\pi: \mathcal{X} \rightarrow \mathbb{R}$:
\begin{talign}
    \Pi[f] = \int_{\mathcal{X}} f(x) \pi(x) dx,
    \label{eq:integration}
\end{talign}
where we assume we have access to evaluations $\{f(x_i)\}_{i=1}^n$ at a set of points $\{x_i\}_{i=1}^n \subseteq \mathcal{X}$.
The problem is particularly challenging if $f$ and $\pi$ are multi-modal and/or are very input-sensitive in different regions of the support.
A plethora of methods exist for tackling this task; the most common are Monte Carlo (MC) methods, which are sampling-based methods that have been studied extensively in theory and practice \citep{Robert2004,Owen2013book}. This subsumes naive Monte Carlo, Markov chain Monte Carlo (MCMC) and quasi-Monte Carlo (QMC).
Sampling is (at least asymptotically, for MCMC) unbiased and thus a gold standard, but precisely for this reason, it can only converge with stochastic rate, and thus requires a large number of samples $n$, both for accuracy and uncertainty quantification.

This is a challenge if evaluations of $f$ or samples from $\pi$ are expensive. 
The former (``expensive $f$'') emerges regularly in climate simulations or other large physical models. Section~\ref{sec:wind_farm} provides an example with a wind farm model -- a field where state-of-the-art models require hundreds of hours of CPU for a single evaluation \citep{Kirby2022,Kirby2023}. The latter (``expensive sampling'') occurs when $\pi$ is a posterior distribution for a complex model conditioned on a large amount of data. Section~\ref{sec:ode_example} illustrates this through an example of Bayesian inference in dynamical system.

In such scenarios, probabilistic numerical methods (PNMs) \citep{Hennig2015,Cockayne2017BPNM,Oates2019Modern,Wenger2021,Hennig2022}, and in particular Bayesian approaches, perform particularly well. For numerical integration, the principle behind Bayesian PNMs is to encode prior information about the integrand $f$, then condition on evaluations of $f$ to obtain a posterior distribution over $\Pi[f]$. 
These methods are well suited for computationally expensive problems since informative priors can be used to encode properties of the problem and to reduce the number of evaluations needed. In addition, the posterior quantifies uncertainty for any finite value of $n$. 

The most popular Bayesian PNM for integration is Bayesian Quadrature (BQ) \citep{ohagan1991bayes,Diaconis1988,Rasmussen2003,briol2019probabilistic}, a method that places a Gaussian Process (GP) \citep{Rasmussen2006} prior on $f$.
With this convenient choice of prior, the posterior on $\Pi[f]$ is a univariate Gaussian, whose mean and variance can be computed in closed form for certain combinations of prior covariance and distribution.  
However, for high-dimensional problems where large amounts of data are necessary, the computational cost of GPs, cubic in $n$, can render BQ too computationally expensive. Fast BQ methods have been proposed to resolve this issue \citep{Karvonen2017symmetric,Jagadeeswaran2018}, but these usually work for a limited range of $\pi$ or $\{x_i\}_{i=1}^n$, and therefore do not provide a widely applicable solution.

This raises the question of whether an alternative probabilistic model could be used in place of a GP within probabilistic integration.
Bayesian neural networks (BNNs) are an obvious candidate, as they are known to work well in high dimensions and with large $n$.
Unfortunately, their application to integration tasks is not straightforward since, in contrast to the GP case, analytical integration of the posterior mean of a BNN is usually intractable. This is a significant challenge which has so far prevented their use for probabilistic numerics.
We resolve this challenge by proposing the concept of \emph{Bayesian Stein (neural) networks} (BSNs), a novel BNN architecture based on a final layer constructed through a Stein operator \citep{Anastasiou2021}. Such choice of architecture is designed specifically so that the resulting BNN is analytically integrable (see Section~\ref{ssec:ls_network}), and hence at our disposal for numerical integration. 

\begin{figure}
  \includegraphics{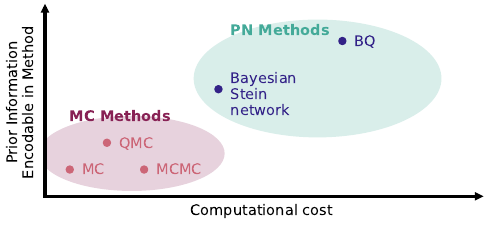}
  \caption{Integration methods can be compared at a high-level in terms of their computational cost and ability to include prior information. In both respects, BSNs provide a compromise in-between MC and BQ.}
  \label{fig:sketch}
\end{figure}

Given these approaches---MC, BQ, BSNs---a natural question remains: ``How should we select a method for a given integration task?''. 
We provide an empirical answer to this question in Section~\ref{sec:experiments}, where we consider a popular benchmark dataset, compute posterior expectations arising in the Bayesian treatment of dynamical systems, and estimate the expected power output of a wind farm. 

Our conclusions are summarized in Figure~\ref{fig:sketch} and presented below. If sampling $\pi$ and evaluating $f$ is computationally cheap, so one can obtain a very large number of data points relative to the complexity of the problem, then MC methods are likely the best choice.
But if $n$ is very limited due to our computational budget, then BQ is likely a better option.
BSNs excel in the intermediate region where $n$ is such that BQ becomes prohibitively expensive but MC is not accurate enough.
The architecture of neural networks, plus sophisticated deep learning software libraries, make training of (small) neural networks memory efficient and fast.
However, achieving good accuracy at low training cost requires special care during training for the Stein architecture. Finding a good training setup is a main contribution of this work, outlined in Section~\ref{sec:methods}. 

For all integration methods, estimates from scarce data are imperfect, so uncertainty estimates are crucial. Bayesian deep learning provides this functionality. Full Bayesian inference is costly even for small neural networks, but we show that a lightweight Laplace approximation \citep{mackay1992practical, ritter2018a} can provide good approximate uncertainty for the Stein network. 

\section{RELATED WORK}
BQ is the method most closely related to our proposed approach and the approach is fully detailed in Appendix~2.
Bayesian PN methods based on alternative priors have also been proposed. These include Bayesian additive regression tree priors \citep{Zhu2020}, multi-output Gaussian process priors \citep{Xi2018MultiOutput,Gessner2019}, and Dirichlet process priors \citep{Oates2017heart}.
These priors each provide different advantages, such as the ability to model highly discontinuous functions, vector-valued functions, or modelling probability distributions respectively. Unfortunately none of these approaches significantly improve scalability, the main goal of our paper. 

The use of (non-Bayesian) neural networks for integration was previously proposed by
\citet{lloyd2020using}. However, their method is only applicable for uniform $\pi$ and shallow networks. \citet{wan2020neural,Si2020} propose to use a Langevin Stein operator applied to a neural network to find good control variates for variance reduction in Monte Carlo approximations (based on an earlier construction by \cite{Oates2017}).
In contrast to their work, we use the neural network to directly compute $\Pi[f]$, and our neural network follows Bayesian semantics and can be used to quantify uncertainty. This requires a different network architecture and an efficient posterior inference algorithm.

\section{BAYESIAN STEIN NETWORKS}

We now describe BSNs. This requires introducing Stein operators, BNNs, and Laplace approximations.
 
\paragraph{Stein Neural Networks}
\label{ssec:stein_operator}
\label{ssec:ls_network}
Stein operators are a technical construction originating in probability theory, but have recently been used as a computational tool \citep{Anastasiou2021}. Building on this line of work, we will use Stein operators to construct the final layer of our BNNs. The reason for this is simple: given some function $u$ (with possibly unknown mean) and a distribution $\pi$, a Stein operator can map $u$ to a mean zero function under $\pi$. This final layer therefore allows us to construct flexible BNNs with the powerful property that \emph{any draw from the posterior will have a known mean under $\pi$}. We now highlight this procedure in detail.

We call $\mathcal{S}$ a \emph{Stein Operator} if for any suitably regular continuously differentiable $u: \mathbb{R}^d \rightarrow \mathbb{R}^d$, the following holds
\begin{talign}
\Pi\left[\mathcal{S}[u]\right] =0.
\label{eq:stein_property}
\end{talign}
Suppose $\mathcal{X} = \mathbb{R}^d$, $\pi$ is continuously differentiable on $\mathcal{X}$, such that $\nabla_x \log \pi$ is well-defined ($[\nabla_x \log \pi(x)]_i = \partial \log \pi(x) / \partial x_i$ for all $i \in \{1,\ldots,d\}$). 
One example of an operator fulfilling \eqref{eq:stein_property} is the diffusion Stein operator \citep{Gorham2016,Barp2019}:
\begin{talign}
   \begin{split}
    \mathcal{S}_m[u](x) & := 
    \begin{aligned}[t]
    &\left(m(x)^\top\nabla_x \log \pi(x)\right)^\top u(x)\\ &+ \nabla_x \cdot \left(m(x)u(x)\right),
    \end{aligned}
    \label{eq:stein_operator}
   \end{split} 
\end{talign}
where $\nabla_x \cdot u(x) = \sum_{i=1}^d \partial u_i(x)/ \partial x_i$, and $m: \mathbb{R}^{d} \rightarrow \mathbb{R}^{d\times d}$ is an invertible matrix-valued function. This operator only requires access to $\nabla_x\log\pi(x_i)$, and can thus be used even if the normalization constant of $\pi$ is unknown.
This is an advantage if $\pi$ is itself a posterior distribution.
In such settings, samples from $\pi$ can be obtained via MCMC, but the distribution $\pi$ itself cannot be evaluated directly.

To construct BSNs, we use an architecture based on a continuously differentiable deep neural network $u_{\theta_u}:\mathcal{X} \rightarrow \mathbb{R}^d$, where $\theta_u \in \Theta_u \subseteq \mathbb{R}^p$, combined with a final layer taking the form of a Stein operator (that we call a \emph{Stein layer}). 
More precisely, we consider an architecture $g_\theta: \mathcal{X} \rightarrow \mathbb{R}$:
\begin{talign}
    g_{\theta}(x) & := \mathcal{S}_m\left[u_{\theta_u}\right](x) + \theta_0.
    \label{eq:stein_network}
\end{talign}
We call this neural network a \emph{Stein neural network} following \citep{wan2020neural,Si2020,Sun2023}, but note that we use the more general diffusion Stein operators $\mathcal{S}_m$ \citep{Gorham2016,Barp2019}. Previous cases can be recovered with $m(x)=I_d$, where $I_d$ is a $d$-dimensional identity matrix, however we will demonstrate in Section~\ref{sec:ode_example} that alternative choices for $m$ can significantly improve the performance of our method.

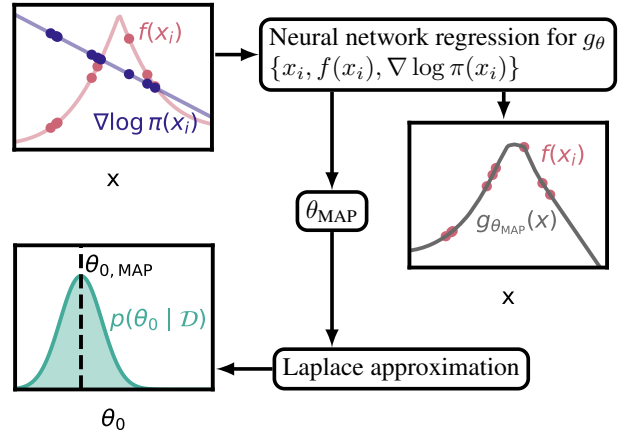
\begin{figure}
    \resizebox{\columnwidth}{!}{\input{tikz/sketch_methods.tikz}}
    \caption[]{\emph{Visualization of BSNs.}
    The BSN prior is conditioned on $\{x_i,f(x_i),\nabla \log \pi(x_i)\}_{i=1}^n$ to obtain a Bayesian posterior on $\theta_0$. This posterior quantifies our uncertainty about $\Pi[f]$. For computational reasons, this posterior is approximated the Laplace approximation around the MAP estimate $\theta_{0, \text{MAP}}$.
    }
    \label{fig:sketch_approach}
\end{figure}

The parameter $\theta = \{\theta_0, \theta_u\} \in \Theta \subseteq \mathbb{R}^{p+1}$ denotes the weights of the neural network $g_\theta$. Thanks to our choice of architecture, \eqref{eq:stein_property} holds and we have:
\begin{talign}
    \Pi \left[g_\theta \right] = \theta_0.
    \label{eq:integral_g}
\end{talign}
The last layer of $g_\theta$ directly tracks the integral of the network, which is the key property for our purpose: by training such a network $g_\theta$ on data from $f$ so that $g_{\theta} \approx f$, we are simultaneously constructing a good approximation of the integral $\Pi[g_\theta] \approx \Pi[f]$ (see Figure~\ref{fig:sketch_approach} for a summary). 

\paragraph{Uncertainty Estimates for Stein Neural Networks} In the context of Bayesian PNM, proposing a BNN architecture is not enough: we are also interested in  \emph{tractable uncertainty estimates over $\Pi[f]$}. We show how to obtain this through the Laplace approximation and a suitable choice of prior, but further details are available in Appendix~3.

The specific architecture of the BSN model means that all the uncertainty on $\Pi[f]$ is represented by the Bayesian posterior on $\theta_0$. This can be obtained through a standard application of Bayes' theorem $p(\theta| \mathcal{D}) \propto p(\mathcal{D}|\theta) p(\theta)$ where in our case the dataset is $\mathcal{D} = \{x_i, f(x_i), \nabla_x \log \pi(x_i) \}_{i=1}^n$, and $p(\theta)$ denotes our prior, $p(\theta| \mathcal{D})$ the posterior and $p(\mathcal{D}|\theta)$ the likelihood. 
The posterior on $\theta_0$ is then the marginal of $p(\theta|\mathcal{D})$. Bayesian inference for deep networks provides uncertainty estimates \citep{neal1996bayesian, mackay1995probable} through $p(\theta| \mathcal{D})$, but this posterior is intractable in general.
MCMC is a prominent tool for approximating $p(\theta| \mathcal{D})$, but using it within an integration method would be circular and re-introduce the spectre of high computational cost \citep{izmailov2021what}.
Other popular approximate inference schemes include variational inference \citep{graves2011practical,blundell2015weight,hinton1993keeping} and ensemble methods \citep{lak2017simple}. Although cheaper, the cost associated with this can still be significant.

We instead opt for the arguably most lightweight approach available for BNNs: the Laplace approximation \citep{mackay1992practical,ritter2018a}.
It is a simple and computationally cheap method, but yet provides competitive uncertainty estimates \citep{daxberger2021laplace}.
The Laplace approximation constructs a second-order Taylor approximation around the mode of the posterior, which amounts to a Gaussian approximate of the posterior around the MAP (maximum a-posteriori) estimate. 
This can be criticized from a Bayesian standpoint as the MAP estimate and the posterior mean of the weights do not necessarily coincide.
However, the MAP estimate is the quantity that is usually tuned in deep learning and is also cheap as it only has to be computed once. 
To be more precise, our approximation of the posterior is implemented in two steps: a Laplace approximation, and an approximation of the corresponding Hessian. 

For the first step, we train the network $g_\theta$ by minimizing the mean squared error loss with weight decay regularizer, given for $\lambda > 0$ by: 
\begin{talign}
    \begin{split}
        l_\text{tot}(\theta) & = l(\theta) + \lambda \|\theta\|_2^2 \\
        \text{where } l(\theta) & = \frac{1}{n} \sum_{i=1}^n \|f(x_i) - g_\theta(x_i)\|_2^2
    \end{split}
\label{eq:loss}
\end{talign}
We notice that $l \propto - \log p(\mathcal{D}| \theta)$ and $\lambda \|\theta\|_2^2 \propto - \log p(\theta)$ whenever we take the prior to be $p(\theta) = \mathcal{N}(\theta \mid 0, \sigma_0^2 I_{p+1})$ ($\sigma_0$ is related to $\lambda$ through a known constant see Appendix~3). As a result, the minimum of the loss above is indeed a MAP estimate: $\theta_{\text{MAP}} = \text{argmin}_\theta l_{\text{tot}}(\theta)$.

Of course, \emph{any} Bayesian treatment of neural networks requires a prior $p(\theta)$. The choice is important since the prior encodes the model class, but there is currently no widely accepted choice. Our choice above was motivated by the fact that for the Laplace approximation, only isotropic Gaussian priors are currently feasible \citep{mackay1992practical,ritter2018a,daxberger2021laplace}. \citet{fortuin2022bayesian} suggest that such priors are undesirable, but \citet{wilson2020bayesian} argue to the contrary: despite their simplicity, such priors still induce sufficiently complex distributions over functions. \citet{daxberger2021laplace} note that it is often beneficial to tune $\sigma_0$ for inference. 

Once the MAP has been identified, we can construct our Laplace approximation using a Taylor approximation (up to second order) of the log-posterior $\log p \left(\theta \mid \mathcal{D}\right)$ around that point. This results in a Gaussian approximation of the posterior distribution: $q_{\text{Laplace}}(\theta) = \mathcal{N}\left(\theta \mid \theta_\text{MAP}, \Sigma \right)$, where $\Sigma$ is proportional to the inverse Hessian of the loss $l_{\text{tot}}$: 
\begin{talign*}
 \Sigma^{-1} &= - \nabla^2 \log p(\mathcal{D}|\theta) - \nabla^2 \log p(\theta)\\
 &= H  +  \sigma_0^{-2} I_{p+1}, \quad \text{where  }
H \propto  \nabla^2_{\theta} l(\theta_\text{MAP})
\end{talign*}
Our second step consists of an approximation of the Hessian. This is necessary since it is often infeasible to calculate $H$ due to the large computational cost when $p$ is large. As a result, we use a positive definite approximation  called the Generalized-Gauss-Newton (GGN; \citep{schraudolph2002fast}) approximation: 
\begin{talign*}
    H_\text{GGN} = \frac{1}{\sigma^2} \sum_{i=1}^n J(x_i) J(x_i)^\top,
\end{talign*}
where $J(x_i) = \nabla_{\theta} g_{\theta}(x_i) |_{\theta=\theta_\text{MAP}}$ and $\sigma$ is the dataset noise. 
This gives us another approximation of the posterior that we denote $q_{\text{GGN-Laplace}}(\theta)$ obtain through $\Sigma^{-1}_{\text{GGN}} = H_{\text{GGN}}  +  \sigma_0^{-2} I_{p+1}$.
Hence, we can extract an approximation of the posterior on the network's prediction of the integral $\Pi[f]$ using Eq. \eqref{eq:integral_g}:
\begin{talign*}
q_{\text{GGN-Laplace}}(\theta_0) = \mathcal{N}\left(\theta_0 | \theta_{0, \text{MAP}}, \left(\Sigma_{\text{GGN}}\right)_{0, 0}\right).
\end{talign*}

\section{Architecture}
\label{sec:methods}

Due to their specific architecture, naive attempts to train BSNs can lead to unsatisfactory results.
Below, as a key contribution, we provide architectural considerations that we have found to significantly improve the conditioning of the loss and lead to better training.

\paragraph*{Choice of Activation Function} We require $u_{\theta_u}$ to be continuously differentiable on $\mathcal{X}$, which imposes restrictions on the activation functions of the BSN. 
A sufficient condition is for these activation functions to be themselves continuously differentiable. 
This excludes the popular RELU activation functions, but includes the CELU (`Continuously Differentiable Exponential Linear Units' \citep{barron2017continuously}; $\text{CELU}(x) = \max(0,x)+ \min(0,\exp(x)-1)$), its continuous extension. It also includes the tanh ($\text{tanh}(x)= (\exp(x)-\exp(-x))/(\exp(x)+\exp(-x))$), Gaussian ($\text{Gauss}(x)=\exp(-x^2)$), and sigmoid ($\text{sigm}(x)=1/(1+\exp(-x))$), TanhShrink  ($\text{TanhShrink}(x)=x-\tanh(x)$) activations. 
We compared activation functions (see Figure~\ref{fig:genz_act} below) and found the CELU to give marginally superior performance on test problems. 
Based on its good performance, we use CELU activations for all experiments.

\paragraph*{Choice of Optimization Procedure}
Optimization for BSNs is challenging due to the unique network architecture.
For one, the architecture contains gradients of the Stein layer, which are harder to train than standard activation functions.
This is because $\nabla_x \log \pi$ can be arbitrarily complicated depending on $\pi$.
We find that the training of $g_\theta$ with Adam \citep{Kingma2015} is considerably slower compared with training $u_{\theta_u}$ (see the Appendix~1.1.1).
We suspect that this is due to the loss landscape of the BSN being more narrow (i.e., having a larger spread in curvature eigenvalue spectrum) than that of $u_{\theta_u}$.
A second order method should alleviate this issue.
Hence, we train the BSN with L-BFGS (an approximate second order method) and the \emph{Hessian-free} optimizer \citep{martens2010deep} (a conjugate gradient based second order method).
And indeed, (approximate) second order optimization reaches much better performance (for an extended discussion see the Appendix~1.1.1).

We therefore used L-BFGS throughout all subsequent experiments. Such quasi-Newton methods have fallen out of fashion in deep learning because they are not stable to noise. In our experiments, we train on the full dataset, so noise is not an issue. We accomplish better (i.e., lower loss) and faster convergence (both in iterations and compute time) with this method compared to gradient descent and its variants.
Note that this approach is only feasible for relatively small (in number of weights $p$) network architectures, as it requires storing the gradient history for the approximate Hessian in memory.
When training on the entire dataset (i.e., no mini-batching), we observe significant speed-up from using GPUs when $n$ is large ($\approx 10^4$).

\paragraph{Choice of $m(x)$}
For most of the experiments we set $m(x) = I_d$, but in general other choices for $m$ are possible.
We test a set of different choices ($m(x) = I_d/(||x||_2^2+1)$, $m(x) = I_d/\sqrt{||x||_2^2+1}$, $m(x) = I_d \pi(x)$, $m(x) = \mathrm{diag}(x)$), but find that none of these perform significantly better than $m(x) = I_d$ (see Appendix~1.1.4 for more details).

\paragraph*{Choice of Point Set} 
BSNs can be implemented regardless of the choice of $\{x_i\}_{i=1}^n$, but we expect better performance when $\{x_i\}_{i=1}^n$ cover regions of high probability under $\pi$. 
A simple solution is to use independent samples from $\pi$; this will be our default choice. 
When independent sampling is not possible, we can use MCMC instead, so long as $\pi$ can be evaluated up to some normalization constant.
Alternatives also include grid of points or QMC point sets (see the Appendix~1.1.2 for a comparison of different point sets), but these are usually only helpful when $\mathcal{X}$ is a hypercube and $\pi$ is uniform. 
Alternatively, one could also use active learning (see \cite{Gunter2014,Briol2015} for corresponding approaches for BQ) based on the Laplace approximation of the uncertainty, but this may not perform well for large $d$, and we did not explore the idea further.

\paragraph{Stein Architecture for Bounded Domains}
The architecture outlined in Section~\ref{ssec:ls_network} is only valid on the open integration domain $\mathcal{X}=\mathbb{R}^d$. 
For bounded $\mathcal{X} \subset \mathbb{R}^d$, it is incorrect because $\Pi[\mathcal{S}_m[u]]=0$ is not necessarily true. 
This can be guaranteed by adding a layer before the Stein layer. 
For example, let $\tilde{u}_{\theta_u}(x) = u_{\theta_u}(x) \delta(x)$, 
where $\delta(x)$ is a smooth function (so that $\tilde{u}_{\theta_u}$ is continuously differentiable) going to zero on the boundary of $\mathcal{X}$. 
Then, $\pi(\cdot)\tilde{u}_{\theta_u}(\cdot)$ is zero on the boundary of $\mathcal{X}$, and as a result $\Pi[\mathcal{S}[\tilde{u}_{\theta_u}]]=0$. 
When $\mathcal{X} = (a, b) \subset \mathbb{R}$, one such function is given by $\delta(x) = (x- a) (b-x)$, and we will use this example where necessary in our experiments.
Beyond bounded $\mathcal{X}$, the architecture can also be adapted to manifold or discrete $\mathcal{X}$; see \cite{Barp2018} and \cite{Shi2022} respectively.  

\section{EXPERIMENTS}
\label{sec:experiments}

\begin{table*}[h!]
  \caption{\emph{Performance on Genz integral family in $d=2$.} Mean relative integration error and standard deviation (based on 5 repetitions) using $n=5120$.} 
  \label{tbl:genz2}
  \begin{center}
  \begin{tabular}{l|lllll}
   ~  &  \multicolumn{3}{c}{Mean Absolute Error} \\
  \textbf{Integrand} &\textbf{MC} &\textbf{BQ} &\textbf{BSN} \\
  \hline 
  Continuous Genz   &1.59e-03 $\pm$ 0.90e-03  &1.40e-03 $\pm$ 0.09e-03          &\textbf{1.11e-05 $\pm$ 0.55e-05}\\
  Discontinuous Genz &2.69e-02 $\pm$ 2.64e-02  &1.12e-02 $\pm$ 0.50e-02          &\textbf{2.56e-03 $\pm$ 1.94e-03}\\
  Gaussian peak     &1.52e-02 $\pm$ 8.85e-03  &\textbf{1.17e-06 $\pm$ 1.11e-06} &1.83e-04 $\pm$ 1.35e-04\\
  Corner peak       &1.85e-02 $\pm$ 1.85e-02  &\textbf{2.49e-04 $\pm$ 1.53e-04} &6.00e-04 $\pm$ 5.39e-04\\
  Oscillatory Genz  &2.88e-01 $\pm$ 1.75e-01  &4.13e-03 $\pm$ 0.89e-03          &\textbf{1.34e-03 $\pm$ 0.97e-03}\\
  Product peak     &7.59e-03 $\pm$ 4.11e-03  &1.82e-04 $\pm$ 0.42e-04          &\textbf{1.42e-04 $\pm$ 0.76e-04}\\
  
  \end{tabular}
  \end{center}
\end{table*}

We consider three main experiments: the  Genz functions benchmark, a parameter inference problem for a dynamical system called Goodwin Oscillator, and an example describing the energy output of a wind farm. 
We compare BSNs to the following approaches:
\begin{itemize}
  \item Monte Carlo methods. When independent sampling from $\pi$ can be used (i.e. for the Genz benchmark and the wind farm experiments) we use MC. When this is not possible, we use instead an MCMC method called Metropolis-Adjusted Langevin algorithm \citep[MALA;][]{roberts1996exponential}. 
  \item A BQ implementation based on \texttt{emukit} \citep{paleyes2019emukit}, with an RBF covariance function $k(x,y)= \lambda\exp(-\|x-y\|^2_2/l^2)$ for some $l,\lambda>0$. 
  We use log-likelihood maximization to choose $l$ and set the GP prior mean to $0$, as we do not have any prior knowledge about the value of the integral.
  In Appendix~1.1.5 we conduct an additional experiment using the Matern 1/2 Kernel.
  However, for this kernel, the posterior mean is only available in $d=1$.
  \item A control functional estimator based on Stein's method (Stein-CF) as described in \cite{oates2019convergence} for the experiments on the Genz dataset and the Goodwin oscillator. The approach can be thought of as a kernel interpolant alternative to our neural network. We use $m(x) = I_d$ and an RBF kernel.
  We use log-likelihood maximization to set the kernel hyperparameters. 
\end{itemize}
To implement the Laplace approximation, we use \texttt{laplace-torch} library \citep{daxberger2021laplace}. 
Across all experiments we employ the same fully connected architecture for $u_{\theta_u}$, where each hidden layer has 32 units, and we use 2 hidden layers (see the Appendix~1.1.3 for more details).

\paragraph{Genz Benchmark} 
\label{sec:genz_exp}

We first consider the Genz family of integrands \citep{genz1984testing}, as a test ground (see Appendix~1.2 for detailed definitions). This benchmark, consisting of six integrands with known integrals, was proposed to highlight the performance of numerical integration methods on challenging tasks including discontinuities, peaks and oscillations. Each integrand has a parameter which can be used to increase the dimensionality $d$ of the domain. We follow the implementation of \citet{Si2020}, where the test functions are transformed to be supported on $\mathcal{X}=\mathbb{R}^d$ and integrated against a multivariate standard Gaussian distribution $\pi$. Since these functions are very cheap, we do not expect BSN or BQ to be competitive with MC methods in terms of runtime, but we use this experiment to showcase the performance of BSNs for challenging integrands and compare methods for fixed $n$.

In Table~\ref{tbl:genz2}, we first consider the case $d=2$ and $n=5120$.
BSN and BQ both outperform MC by several orders of magnitude in terms of mean relative integration error.
Notably, BSN is significantly better than BQ for the discontinuous Genz function, indicating that the neural network is able to adapt to rapidly changing functions.
For the Gaussian Genz function, BQ outperforms the BSN due to the fact that the prior is more informative.
Both methods lead to a significant improvement over MC, but we can run the BSN at higher number of data points $n$ than BQ (see Appendix~1.2).

\begin{figure}[h]
  \includegraphics{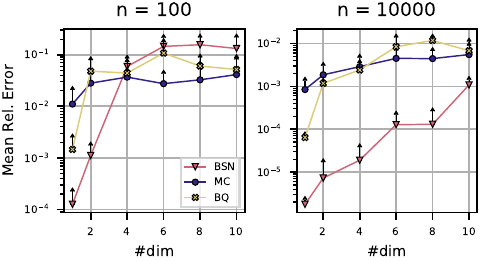}
  \caption{\emph{Continuous Genz function}. We compare methods as a function of $d$ for $n=100$ (\emph{left}) and $n=10000$ (\emph{right})(mean and standard deviation based on 5 repetitions).}
  \label{fig:genz_dim_main}
\end{figure}
We then considered the impact of dimensionality on MC, BQ, and BSN in Figure~\ref{fig:genz_dim_main}. We focus on the Continuous Genz function for simplicity.
If too few evaluations $n$ are available, the Stein network cannot approximate $f$ well, but with a sufficiently large $n$ (i.e. $n \approx 10^2$ in $d=1$ and $n \approx 10^4$ in $d=10$), BSN significantly outperforms MC and BQ.

\begin{figure}[h]
  \includegraphics[]{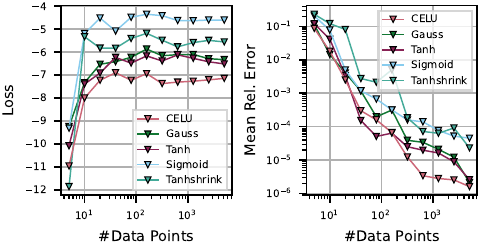}
  \caption{\emph{Impact of the choice of activation function for the Continuous Genz function.}
  Loss $l$ (\emph{left}) and mean relative integration error (\emph{right}) (mean based on 5 repetitions) as a function of $n$.}
  \label{fig:genz_act}
\end{figure}

We also considered the impact of the choice of activation functions for $u_{\theta_u}$ in Figure~\ref{fig:genz_act}. Again, we focus on the Continuous Genz integrand, but limit ourselves to $d=1$.
We consider a diverse set of activation functions (described in Section~\ref{sec:methods}), all continuously differentiable as required for the final Stein layer. We find that the CELU activation leads to the best results on the Continuous Genz dataset, but other activation functions like the tanh and Gaussian activations also perform well.

\begin{figure*}[h]
	\includegraphics{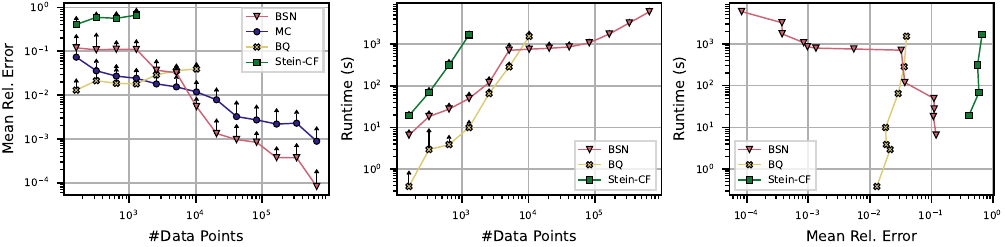}
	\caption{\emph{Continuous Genz function in $d=20$}. 
  Mean relative integration error (\emph{left}), and run time (\emph{center}) (mean and standard deviation based on 5 repetitions) as a function of $n$. 
  \emph{Right:} Run time in seconds as a function of mean relative integration error.
  }
	\label{fig:genz_20}
\end{figure*}

Finally, we have a deeper look at the Continuous Genz function in $d=20$ in Figure~\ref{fig:genz_20}. 
We observe that a large enough $n$ ($n \approx 10^4$) is necessary for the interpolation capabilities of the model to significantly improve performance.
In those cases, the BSN achieves significantly better performance than MC-sampling.
We note that MC sampling is cheap on the Genz benchmark dataset, and this benchmark is only used as a test bed to vary the complexity of our integrands, so we only compare the MC to the other methods in terms of sample efficiency.
Both BQ and Stein-CF do not achieve good performance and are too expensive (in runtime and in memory) to run for large $n$.
The BSN can perform well even for much larger datasets (we ran it up to $n \approx 10^6$).

\begin{figure}
\includegraphics{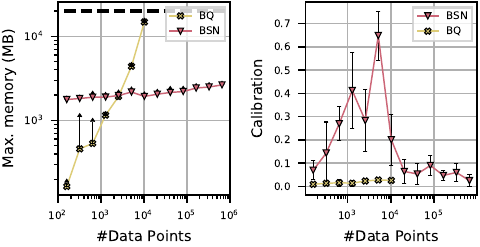}
\caption{\emph{Continuous Genz function in $d=20$}.
Memory requirements (\emph{left}) and calibration (\emph{right}) (mean and standard deviation based on 5 repetitions) as a function of $n$. 
}
\label{fig:memory}
\label{fig:genz_20_2}
\end{figure}

We can compare the BSN and BQ not only in runtime but also in terms of memory requirements.
However, computing accurate memory requirements in python can be difficult as common python libraries use for example \texttt{C++} backends.
The memory requirements of these non-python backends is commonly not taken into account using the built-in memory profiler.
So instead, we use the profiler of our cluster, which outputs the maximum memory required by the program.
Figure~\ref{fig:memory} shows that the BSN memory requirements increase more slowly than for BQ.
Both kernel based methods (BQ and CF) surpass our allotted memory limit of 20~GB.

To evaluate the uncertainty estimates provided by the GGN-Laplace approach, we calculate their calibration $\gamma$.
The calibration is given by the ratio between relative integration error $e_{\text{abs}}$, and the standard deviation $\sigma_{\theta_0}$ of the GGN-Laplace approximation of the posterior on $\theta_0$: $\gamma = e_{\text{abs}}/\sigma_{\theta_0}$. Similarly, for BQ, $\sigma_{\theta_0}$ is the posterior standard deviation on $\Pi[f]$.
A calibration fluctuating around one indicates a well calibrated model, and a large calibration suggests a model that is overconfident, rendering its uncertainty estimates unreliable.
The GGN-Laplace approach as well as BQ lead to uncertainty estimates which are underconfident (although less so for the BSN), especially in the high data regime (see Figure~\ref{fig:genz_20_2}).
Underconfident predictions are still useful in that they provide a prudent assessment of our uncertainty.

\paragraph{Bayesian Inference for the Goodwin Oscillator}
\label{sec:ode_example}
A challenging computational task in Bayesian inference is posterior inference for parameters of dynamical systems (see for example \cite{Calderhead2011}). The challenge is due to the large computational cost of posterior sampling, which is incurred due to the need to solve systems of differential equations numerically at a high-level of accuracy. In addition, large datasets can further increase the computational cost, making the task a prime candidate for BSNs. 
For this experiment, we consider parameter inference in a dynamical system called the Goodwin oscillator \citep{goodwin1965oscillatory}. This model describes how the feedback loop between mRNA transcription and protein expression can lead to oscillatory dynamics in a cell. It is a common benchmark for MC methods \citep{Calderhead2009,Oates2016thermo,Riabiz2020}.

We analyse the setting with no intermediate protein species, leading to a system with $d=4$ parameters: $x = (a_1, a_2, k, \alpha) \in \mathbb{R}_{+}^4$.
Given a posterior distribution $\pi$, we want to compute the posterior mean $\Pi[f]$ of each of the ODE parameters, i.e., $f(x)= x$. 
For this experiment, the posterior distribution is conditioned on a synthetic dataset of $2400$ observations generated for some known parameter values. 
Our exact experimental setup is based on \citep{chen2019stein}, and we refer to the Appendix~1.3 for a detailed description.

The posterior density $\pi$ is only available in unnormalized form, and we therefore use MALA for sampling.  This is relatively expensive: sampling $n=1000$ realizations takes around $30$ seconds, which is on the same timescale as network inference ($\sim 1$ min).
For ODE problems requiring more complex solvers or settings with a large dataset, the sampling time might increase even further. 

\begin{figure}[]
  \includegraphics{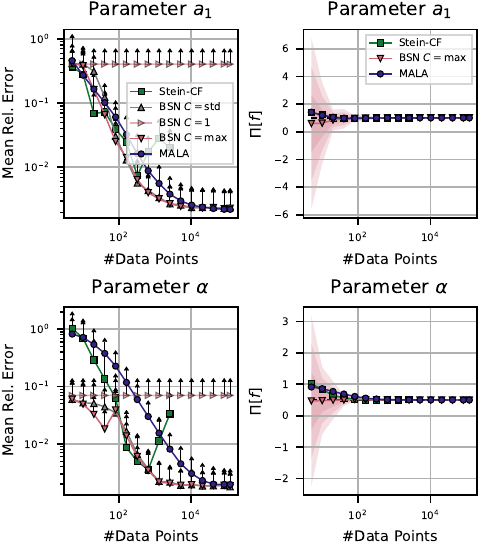}
  \caption{
  \emph{Posterior expectations for the parameters of a Goodwin ODE}.
  Mean relative integration error and standard deviation (\emph{top-left} and \emph{bottom-left}), and uncertainty estimates (\emph{top-right} and \emph{bottom-right}) (based on 5 repetitions) as a function of $n$. 
  }
  \label{fig:goodwin_oscillator}
\end{figure}

In this setting, $\nabla_x \log \pi(x)$ can take very large values, which makes training the BSN harder. 
We find that $m(x) = I_d/C$ for $C \in \mathbb{R}$ can considerably improve the performance. We considered two choices for the constant $C$:
\begin{itemize}
  \item the standard deviation of $\{\nabla_x \log \pi(x_i)\}_{i=1}^n$ (called $C = \text{std}$ in Figure~\ref{fig:goodwin_oscillator}). 
  \item the largest score value: $C = \max_{i=1,\ldots,n} \nabla_x \log \pi(x_i)$ ($C=\text{max}$ in Figure~\ref{fig:goodwin_oscillator}). 
\end{itemize}
Figure~\ref{fig:goodwin_oscillator} compares the performance of the proposed regularizations.
Both choices work well, in contrast to using no regularization at all (i.e. $C=1$). 
We find that the BSN either matches the performance of MALA (for parameter $\alpha$) or surpasses it (parameter $a_1$).
The Stein-CF performs well but struggles in the high data regime due to unstable hyperparameter optimization.
The results for $a_2$ and $k$ are presented in Appendix~1.3. 
The saturation in reached accuracy for both the BSN and MALA can be attributed to the noisy likelihood evaluations. 

Before concluding, we emphasize that BSN is the only available Bayesian PNM here. This is because $\pi$ is unnormalized and BQ 
\update{
  would require an additional step of computing the normalization constant, which would lead to additional runtime and likely incur additional numerical error.
}

\paragraph{Expected Local Turbine Thrust Coefficient for Wind Farm Layout Design}
\label{sec:wind_farm}
\begin{figure*}[h]
  \includegraphics{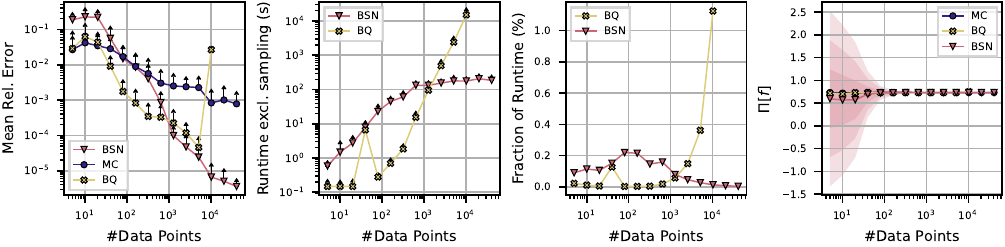}
  \caption{\emph{Wind farm model}. Mean relative integration error (\emph{left}), and run time (\emph{centre-left}) (mean and standard deviation based on 5 repetitions) as a function of $n$. 
  \emph{Center-right:} Fraction of runtime BSN and BQ contribute to the total runtime which includes the runtime of the wind farm simulation.
  \emph{Right:} Uncertainty estimates provided by the Laplace approximation.
  }
  \label{fig:wind_farm_7d}
\end{figure*}
The energy produced by a wind farm depends on factors including the distance between turbines, the direction of the wind, and the wake produced by individual turbines. 
To understand this phenomenon, fluid dynamic simulations can be used to estimate a local turbine thrust coefficient (which we denote $f$), which largely determines energy production \citep{Nishino2016}. Since a number of these factors are unknown, it is common practice to represent uncertainty through a distribution (denoted $\pi$), and calculate the \emph{expected} local turbine thrust coefficient $\Pi[f]$.

A particular challenge here is the cost of evaluating $f$. For the model we are using (a low-order wake model from \cite{Niayifar2016}), each evaluation of $f$ takes approximately $130$ seconds, but more accurate models \citep{Kirby2022} can take up to tens of hours per evaluation. However, it is well known that $f$ is a smooth function of the inputs, which makes Bayesian PNMs, such as BSNs, prime candidates for the task.

The input to our model $f$ are the wind direction, the turbulence intensity, as well as a number of parameters representing the design of the wind farm (including parameters impacting the distance between turbines, and turbine-specific parameters such as the turbine resistance coefficient, the turbine hub height and diameter, and parameters describe the turbine wake). 
The distribution $\pi$ consists of independent distributions (either mixtures of Gaussians, or a truncated Gaussian) on each input to the wake model. 
The Appendix~1.4 provides full details on the wind farm dataset.

The results are presented in Figure~\ref{fig:wind_farm_7d}. Since the ground truth is unknown for this problem, we ran BSN on a dataset which is $5$ times larger than what is plotted in order to get a benchmark value. We compared the runtime of all methods including sampling, where we assume that all the points were sampled sequentially (corresponding to running the experiment on a single CPU).
The additional runtime of both BQ and the BSN is negligible compared to the initial sampling time.
Both methods achieve a much lower mean relative integration error compared to sampling, clearly demonstrating the power of Bayesian PNM methods for problems involving expensive integrands.

On this dataset BQ cannot be used to compute uncertainty estimates, because we cannot integrate the kernel twice in closed form for truncated Gaussians.
However, the uncertainty estimates computed with the Laplace approximation for the BSN accurately capture deviations from the ground truth value (shown in Figure~\ref{fig:wind_farm_7d}).

\section{LIMITATIONS AND DISCUSSION}

The primary advantage of BSNs is in terms of scalability, but they also suffer from some limitations, discussed below.

Firstly, in contrast to GPs where prior knowledge (such as periodicity or smoothness) about $f$ can be encoded via a kernel, selecting good functional priors for BNNs can be challenging.
\update{
In general, the BSN encodes that the integrand $f$ is smooth, as the outputs of the BSN are smooth functions (the BSN is composition of smooth functions).
}
Our experiments show that simple prior choices are often sufficient to achieve good results for moderately hard problems. More advanced options \citep{sun2018functional,Pearce2019} could be considered, but this would require novel Laplace approximations.
\update{
In cases the BNN prior is misspecified (i.e., the function $f$ is not contained in the prior), we can always increase the number of parameters of the neural network.
}

Secondly, our experiments suggest convergence with large $n$. Although we did not analyse this convergence from a theoretical viewpoint, we note that \citet[Proposition 1 and 2]{Si2020} can be used to prove consistency of the BSN posterior mean to the true value of the integral. 
Currently, we do not have any results for the convergence rates, but this could be an interesting direction for future research (for example, \citet{Belomestny2023} provides a rate for a related approach).
This is in contrast with the GP case where convergence results are highly developed \citep{Kanagawa2017,Kanagawa2019,Karvonen2020,Wynne2020}.

Thirdly, computational cost is highly dependent on the complexity of the deep network $u_{\theta_u}$. Using standard matrix multiplication, neural network training is linear in the number of (non-bias) parameters $p$, the number of training samples $n$, and the number of gradient iterations $i$, i.e, $O(p n i)$. 
Across all our experiments we used the same architecture for $u_{\theta_u}$ independent of $n$, but we expect that the complexity of the network will need to increase significantly when high accuracy is required. In such cases, we expect that mini-batching and first order optimization could improve \emph{scalability}, but would likely incur new issues with \emph{stability}.

\section{CONCLUSION}

We have introduced a way to leverage the function approximation abilities of deep BNNs specifically for integration through the application of a Stein operator. 
Employing a Laplace approximation provides uncertainty quantification of good quality in this architecture. 
We have noted that significant work is required to stabilize the training process to this end: both the architecture and the training method must be adapted to the non-standard form of the loss.

BSNs perform consistently well across experiments, both in accuracy and in runtime, and are thus an interesting alternative to BQ, especially for the intermediate regime between very small sample size (where traditional BQ works well), and very large sample numbers (where classic MC methods continue to be the preferred solution).
Our experiments on a variety of applications also highlight some functional strengths of the BSN approach. In particular, it can deal flexibly with a wide range of integration densities, including cases in which the density is known in unnormalized form.

\begin{acknowledgements}
The authors would like to thank Zhuo Sun, Lukas Tatzel, Frank Schneider and Andrew Kirby for helpful discussions. We would also like to thank Andrew Kirby for sharing code to run the wind-farm experiments, which is available at \url{https://github.com/AndrewKirby2/ctstar_statistical_model}. 
Part of this work was initiated by Dagstuhl Seminar 21432 "Probabilistic Numerical Methods - From Theory to Implementation."
PH gratefully acknowledges financial support by the European Research Council through ERC StG Action 757275 / PANAMA; the DFG Cluster of Excellence “Machine Learning - New Perspectives for Science”, EXC 2064/1, project number 390727645; the German Federal Ministry of Education and Research (BMBF) through the Tübingen AI Center (FKZ: 01IS18039A); and funds from the Ministry of Science, Research and Arts of the State of Baden-Württemberg. FXB was supported by the Lloyd's Register Foundation Programme on Data-Centric Engineering and The Alan Turing Institute under the EPSRC grant [EP/N510129/1], and through an Amazon Research Award on “Transfer Learning for Numerical Integration in Expensive Machine Learning Systems”. 
\end{acknowledgements}

\bibliography{sample}
\onecolumn 
\begin{appendices}
\section{EXPERIMENTAL DETAILS}

Below we give implementation details for all the datasets used and provide additional experimental results.

For our implementation of the BSN and the experiments described in the main text we make use of the following packages: \texttt{PyTorch} \citep{pytorch}, \texttt{emukit} \citep{paleyes2019emukit}, \texttt{GPyTorch} \citep{gpytorch}, \texttt{laplace-torch} \citep{daxberger2021laplace}, \texttt{PyWake}\citep{pywake}, and \texttt{Matplotlib} \citep{matplotlib}.

\subsection{Impact of Architecture Design}
\label{sec:sup_architecture}

We provide additional discussion concerning the choice of activation function, choice of sampling strategy and choice of optimizer.

\subsubsection{Choice of Optimizer}
\label{sec:optimizer}

We compare different optimizers for the BSN and a standard neural network, where for the standard neural network we use the same architecture as for $u_{\theta_u}$.
We use the 1-dimensional wind farm dataset with $n=320$ data points.
We choose this dataset, due to the complicated structure of the score function of a mixture of Gaussians.
For Adam, we use mini-batching with a batch size of $32$ and full-batch training.
For the Hessian-free optimizer and L-BFGS we only consider full-batch training.
For Adam, we use $10000$ iterations, for the Hessian-free optimizer $1000$ iterations, and for L-BFGS we use automatic stopping based on the strong Wolfe conditions. 
We compare the loss for all training methods.
We also use CELU and RELU activation functions, where RELU is included as it is the standard activation function for neural networks.

\begin{figure}
\includegraphics{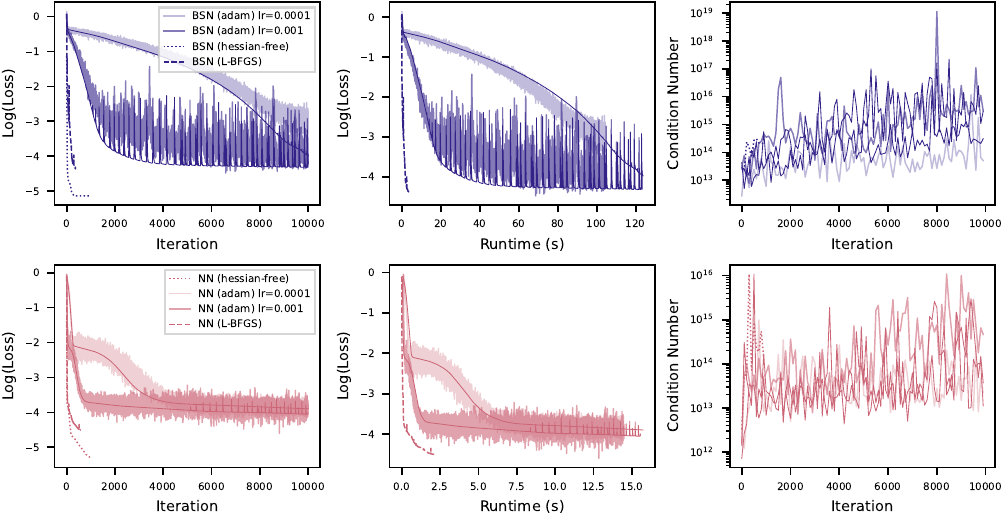}
\caption{Regression performance of a plain neural network (red) and a BSN (blue) using CELU activations. 
Loss (\emph{left}), and condition number (\emph{left}) as a function of the iteration. 
\emph{Centre:} loss as a function of the runtime.
Thin dark lines correspond to training with full-batch Adam.
Runtime of the Hessian-free optimizer not plotted, due to its long runtime.
}
\label{fig:regression_performance}
\end{figure}

\begin{figure}
\includegraphics{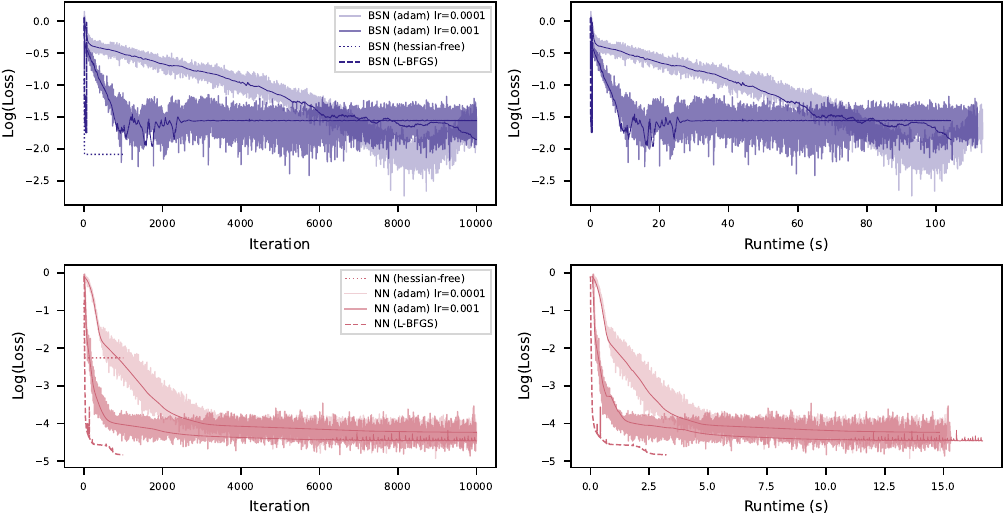}
\caption{Regression performance of a plain neural network (red) and a BSN (blue) using RELU activations. 
Loss (\emph{left}) as a function of the iteration. 
\emph{Centre:} loss as a function of the runtime.
Thin dark lines correspond to training with full-batch Adam.
Runtime of the Hessian-free optimizer not plotted, due to its long runtime.
}
\label{fig:regression_performance_relu}
\end{figure}

Training a standard neural network with RELUs work significantly better, than using CELUs both in terms of the loss reached at the end of training and in terms of runtime (see Figure~\ref{fig:regression_performance} and Figure~\ref{fig:regression_performance_relu}).
Using RELUs does not work for the BSN, as the gradients of $u_{\theta_u}$ lead to discontinuities.

Training of the BSN using Adam is considerably slower than the training progress of the standard neural network.
We find that for CELU activation function, using (approximate) second order methods leads to a large improvement both in terms of speed and loss.
The success of the second order methods might be due to a narrow loss landscape, i.e., a larger spread in the eigenvalue spectrum of the curvature.
Therefore, we also examine the condition number of the Hessian, and we find that the BSN has a slightly higher condition number than the standard neural network (we do not report the condition number for RELUs as it cannot be computed numerically).
Given its short runtime and good optimization results, we choose L-BFGS for all our experiments.

\subsubsection{Sampling Strategies}
\label{sec:sup_sampling}

\begin{figure}
\includegraphics{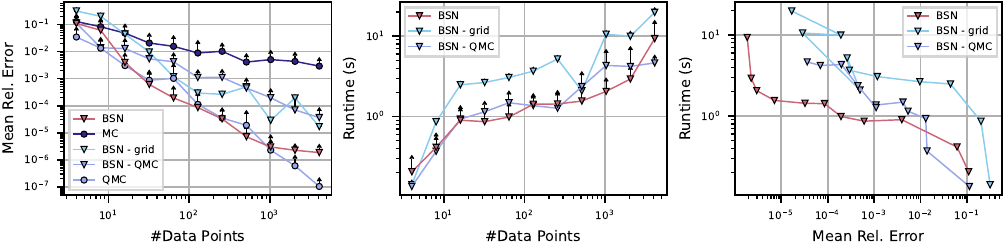}
\caption{Comparing different sampling schemes on the continuous Genz dataset in $d=1$. The BSN is trained on MC-sampled points, QMC-sampled points and on a regular grid.
  Mean relative integration error (\emph{left}), and runtime (\emph{centre}), (based on 5 repetitions) as a function of $n$. 
  \emph{Right:} Run time in seconds as a function of mean relative integration error.
}
\label{fig:genz_1_sampling}
\end{figure}

For our experiments in the main text, we choose the data points by sampling from $\pi$, i.e., $x_i \sim \pi$.
Here we consider two additional sampling strategies:
\begin{itemize}
  \item Using a quasi-Monte Carlo (QMC) sequence. We use \texttt{SciPy}'s \citep{scipy} implementation of QMC based on the Sobol sequence \citep{sobol1967on}.
  \item Linearly spaced points in a hypercube (called \emph{grid} in Figure~\ref{fig:genz_1_sampling}). Here we consider the hypercube $[-5 \sigma_{\pi}, 5 \sigma_{\pi}]^d$, where $\pi(x) = \mathcal{N}(x | 0, \sigma_{\pi})$.
\end{itemize}

Figure~\ref{fig:genz_1_sampling} shows the result of the different sampling strategies in $d=1$.
The BSN performs better using MC samples then using QMC samples and grid points.
The low performance of the latter is expected, since too few points a placed in regions with a high probability mass.

\subsubsection{Choice of Architecture}
\label{sec:architecture_search}

\begin{figure}
\includegraphics{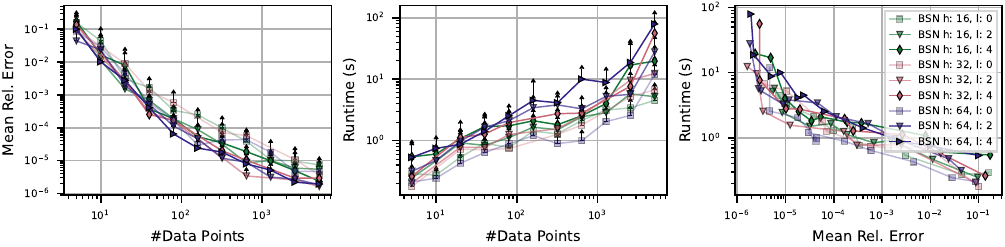}
\caption{
  Testing different architectures on the 1-dimensional continuous Genz dataset.
  Mean relative integration error (\emph{left}), and run time (\emph{centre}), (based on 5 repetitions) as a function of $n$. 
  \emph{Right:} Run time in seconds as a function of mean relative integration error.
}
\label{fig:architecture_search}
\end{figure}

We consider a basic architecture of the following form:
\begin{talign*}
    u_{\theta_u} = \text{Linear}(d, h) \circ CELU (\circ \text{Linear}(h, h) \circ CELU)^l \circ \text{Linear}(h, d), 
\end{talign*}
where $h$ are the number of hidden units and $l$ are the number of hidden layers.
Figure~\ref{fig:architecture_search} shows the performance of different architectures on the 1-dimensional continuous Genz dataset.
All architectures perform similar but the architecture with $l=2$ and $h=32$ reaches the lowest error the fastest for large $n$.
Hence, we use this architecture for our experiments.

\begin{figure}
\includegraphics{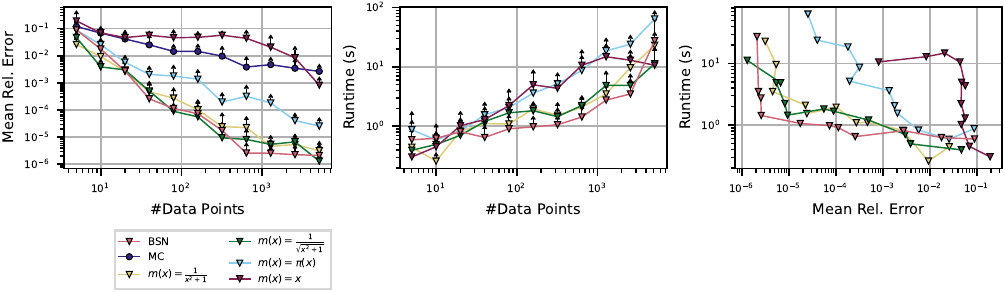}
\caption{\emph{Continuous Genz dataset in $d=1$ with different $m(x)$.}
Mean relative integration error (\emph{left}), run time (\emph{center}) (based on 5 repetitions) as a function of $n$. 
\emph{Right:} Run time in seconds as a function of mean relative integration error.
}
\label{fig:genz_m}
\end{figure}
\subsubsection{Choice of $m(x)$}
For most of our experiments we set $m(x) = I_{d}$ in Equation~\eqref{eq:stein_operator}.
This might not necessarily be the best choice for a given task, but finding a function $m$ that works well is hard.
We test different $m$ on the 1-dimensional Continuous Genz function:
\begin{itemize}
  \item $m(x) = \frac{I_d}{||x||_2^2 +1}$ - $m(x)$ goes to zero for $x \rightarrow \pm\infty$
  \item $m(x) = \frac{I_d}{\sqrt{||x||_2^2 +1}}$ - $m(x)$ goes to zero for $x \rightarrow \pm\infty$ and cancels the $\nabla_x \log \pi(x)$ term for large $x$.
  \item $m(x) = I_d \pi(x)$ - in cases where $\pi$ is a normal distribution, this function also goes to zero for $x \rightarrow \pm\infty$.
  \item $m(x) = \mathrm{diag} x$ - example of a function having negative effect.
\end{itemize} 
The results of comparing these different $m$ are shown in Figure~\ref{fig:genz_m}.
On this test problem, none of the proposed $m$ significantly outperforms the choice $m(x)=I_d$, with some performing significantly worse.

\subsubsection{Choice of GP Kernel}

\begin{figure}
\includegraphics{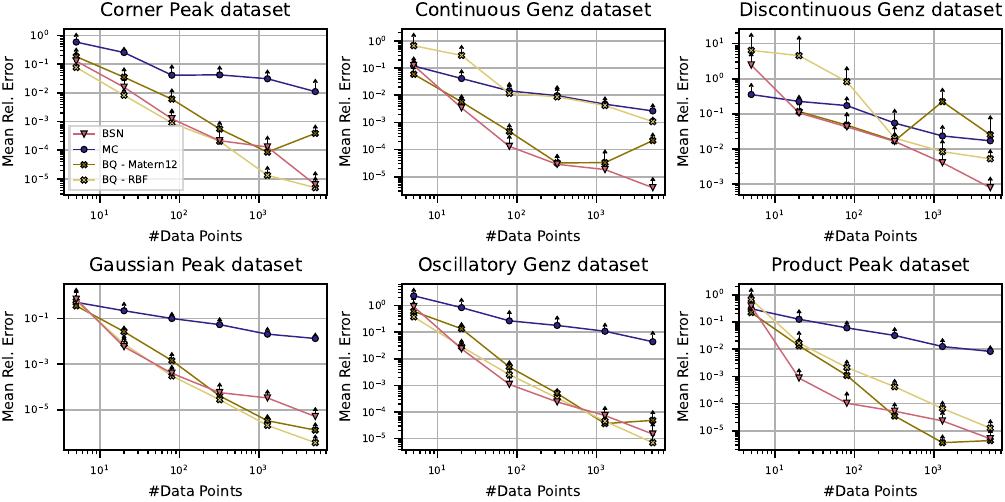}
\caption{\emph{BQ with Matern 1/2 kernel on the Genz family in $d=1$.}
Mean relative integration error (based on 5 repetitions) as a function of $n$. 
}
\label{fig:genz_matern}
\end{figure}

  As a benchmark we use BQ with an RBF kernel for all our experiments.
  The reason for this choice of kernel is the closed form availability of posterior mean and covariance when $\pi$ is a normal distribution.
  Here we add an experiment using a Matern 1/2 kernel.
  For this choice of kernel the posterior mean is only available in $d=1$, hence we conduct the experiment on the 1 dimensional Genz dataset (see Section~\ref{sec:matern} for the expression of the kernel mean embedding).
  The corresponding results a found in Figure~\ref{fig:genz_matern}. Once again, we do not observe a significant difference in performance, except for the continuous Genz dataset. 

\subsection{Genz Benchmark}
\label{sec:genz_family}
\label{sec:exp_details_genz}
In our experiments we use the Genz integrand family dataset.
Here we include a short description of each dataset, plus additional experiments on the 2-dimensional version of each dataset.
In our experiments we integrate the Genz function against a standard normal $\pi(x) = \mathcal{N}(x | 0, 1)$.
This requires the transformation of the inputs to the original Genz functions $f$, which are to be integrated against $[0, 1]^d$.
Therefore, we compute $\Pi_{\pi}[f \circ c]$ where $c(x) = \frac{1}{2} \left( 1 + \text{erf} \left( \frac{x}{\sqrt{2}} \right) \right)$ is the cumulative density function of the standard normal.
We give the form of $f$ below.

\begin{figure}
\includegraphics{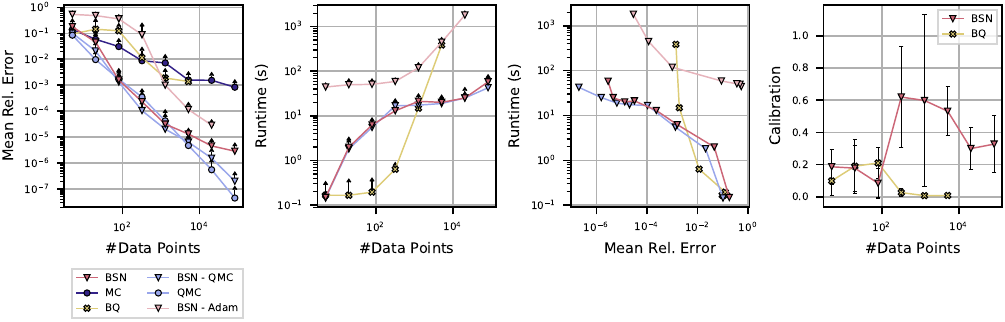}
\caption{\emph{Continuous Genz dataset in $d=2$.}
Mean relative integration error (\emph{left}), run time (\emph{centre-left}), and calibration (\emph{right}) (based on 5 repetitions) as a function of $n$. 
\emph{Center-right:} Run time in seconds as a function of mean relative integration error.
}
\label{fig:genz_continuous}
\end{figure}

\paragraph{Continuous Genz dataset}
The integrand is given by
\begin{talign*}
  f(x) = \exp \left( - \sum_{k=1}^d a_k | x_k - u_k| \right)
\end{talign*}
with parameters $a_k = 1.3$ and $u_k = 0.55$.
See Figure~\ref{fig:genz_continuous} for results on a 2 dimensional version of this dataset.

\begin{figure}
\includegraphics{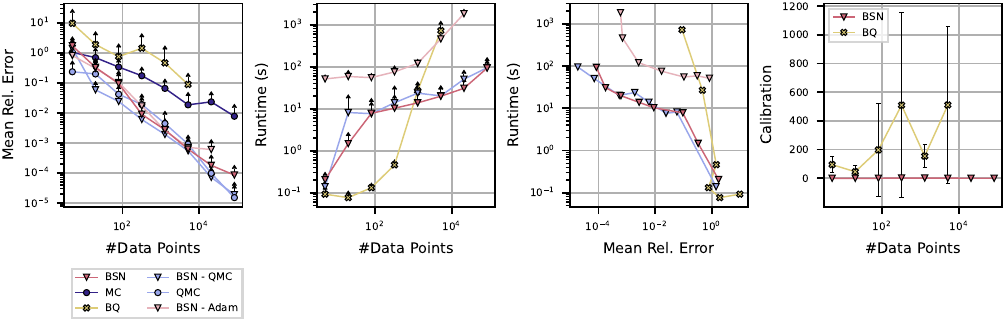}
\caption{\emph{Corner Peak dataset in $d=2$.}
Mean relative integration error (\emph{left}), run time (\emph{centre-left}), and calibration (\emph{right}) (based on 5 repetitions) as a function of $n$. 
\emph{Center-right:} Run time in seconds as a function of mean relative integration error.
}
\label{fig:genz_corner}
\end{figure}

\paragraph{Corner Peak dataset}
The integrand is given by
\begin{talign*}
  f(x) = \left(1 +  \sum_{k=1}^d a_k x_k \right)^{-(d+1)}
\end{talign*}
with parameters $a_k = 5$.
See Figure~\ref{fig:genz_corner} for results on a 2 dimensional version of this dataset.

\begin{figure}
\includegraphics{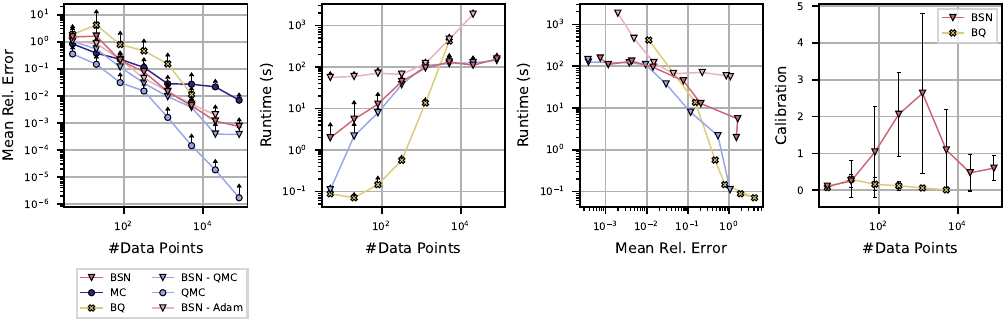}
\caption{\emph{Discontinuous Genz dataset in $d=2$.}
Mean relative integration error (\emph{left}), run time (\emph{centre-left}), and calibration (\emph{right}) (based on 5 repetitions) as a function of $n$. 
\emph{Center-right:} Run time in seconds as a function of mean relative integration error.
}
\label{fig:genz_discontinuous}
\end{figure}

\paragraph{Discontinuous Genz dataset}
The integrand is given by
\begin{talign*}
  f(x) = \begin{cases}
    0, \quad\text{if}~ x_k > u_k ~\text{for any}~ $k$\\
    \exp\left( \sum_{k=1}^d a_k x_k \right)
  \end{cases}
\end{talign*}
with parameters $a_k = 5$ and $u_k=0.5$.
See Figure~\ref{fig:genz_discontinuous} for results on a 2 dimensional version of this dataset.

\begin{figure}
\includegraphics{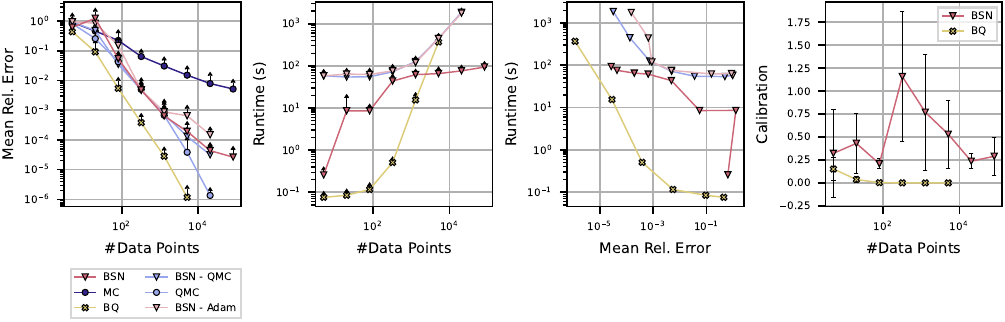}
\caption{\emph{Gaussian peak dataset in $d=2$.}
Mean relative integration error (\emph{left}), run time (\emph{centre-left}), and calibration (\emph{right}) (based on 5 repetitions) as a function of $n$. 
\emph{Center-right:} Run time in seconds as a function of mean relative integration error.
}
\label{fig:genz_gaussian}
\end{figure}

\paragraph{Gaussian peak dataset}
The integrand is given by
\begin{talign*}
  f(x) = \exp \left( - \sum_{k=1}^d a_k^2 (x_k - u_k)^2 \right)
\end{talign*}
with parameters $a_k = 5$ and $u_k = 0.5$
See Figure~\ref{fig:genz_gaussian} for results on a 2 dimensional version of this dataset.

\begin{figure}
\includegraphics{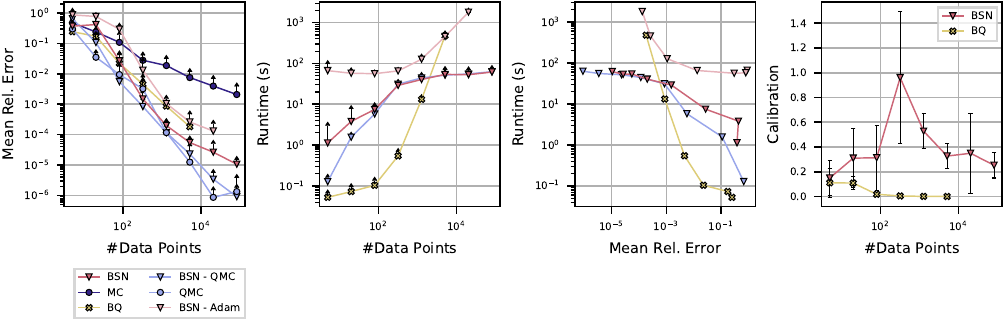}
\caption{\emph{Product peak dataset in $d=2$.}
Mean relative integration error (\emph{left}), run time (\emph{centre-left}), and calibration (\emph{right}) (based on 5 repetitions) as a function of $n$. 
\emph{Center-right:} Run time in seconds as a function of mean relative integration error.
}
\label{fig:genz_product}
\end{figure}

\paragraph{Product peak dataset}
The integrand is given by
\begin{talign*}
  f(x) = \prod_{k=1}^d \frac{1}{\left( a_k^{-2} + (x_k - u_k)^2\right)}
\end{talign*}
with parameters $a_k = 5$ and $u_k = 0.5$
See Figure~\ref{fig:genz_product} for results on a 2 dimensional version of this dataset.

\begin{figure}
\includegraphics{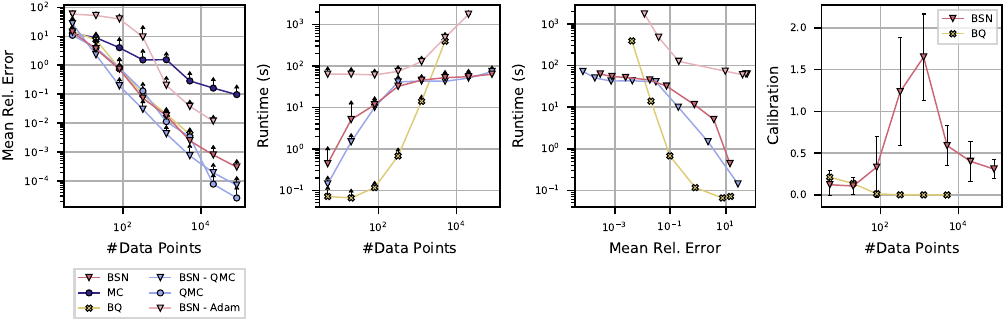}
\caption{
\emph{Oscillatory Genz dataset in $d=2$.}
Mean relative integration error (\emph{left}), run time (\emph{centre-left}), and calibration (\emph{right}) (based on 5 repetitions) as a function of $n$. 
\emph{Center-right:} Run time in seconds as a function of mean relative integration error.
  }
\label{fig:genz_oscillatory}
\end{figure}

\paragraph{Oscillatory Genz dataset}
The integrand is given by
\begin{talign*}
  f(x) = \cos \left(2 \pi u + \sum_{k+1}^d a_k x_k \right)
\end{talign*}
with parameters $a_k = 5$ and $u = 0.5$
See Figure~\ref{fig:genz_oscillatory} for results on a 2 dimensional version of this dataset.

\subsection{Goodwin Oscillator}
\label{sec:goodwin_oscillator}

Goodwin oscillator \citep{goodwin1965oscillatory} describes how the feedback loop between mRNA transcription and protein expression can lead to oscillatory dynamics in a cell. 
We here consider the case with no intermediate protein species.
The experimental setup is based on earlier work by \citep{Riabiz2020, chen2019stein, Calderhead2009, Oates2016thermo}.

The Goodwin oscillator with no intermediate protein species is given by:
\begin{talign*}
  \frac{du_1}{dt} &= \frac{a_1}{1 + a_2 u_2^\rho} - \alpha u_1\\
  \frac{du_2}{dt} &= k_1 u_1 - \alpha u_2,
\end{talign*}
where $u_1$ corresponds to the concentration of mRNA and $u_2$ to the concentration of the corresponding protein product.
We set $\rho = 10$.

As initial conditions we set $u_0 = (0, 0)$.
To generate the ground truth dataset, we set $a_1 = 1$, $a_2 = 3$, $k_1=1$ and $\alpha=0.5$.
We use a measurement noise of $\sigma = (0.1, 0.05)$.
Data was collected for $2400$ time points in $t\in[1, 25]$, leading to the following expression for the likelihood:
\begin{talign*}
 p(y | x) \propto \exp \left(- \frac{1}{2 \sigma_1^2} \sum_{k=1}^{2400} ||y_{1,k} - u_1(t_k) ||_2^2- \frac{1}{2 \sigma_2^2} \sum_{k=1}^{2400} ||y_{2,k} - u_2(t_k) ||_2^2 \right)
\end{talign*}
We use an \texttt{JAX}'s implementation of Dopri5(4) to solve the ODE. 
We use automatic differentiation implemented in \texttt{JAX} to compute derivatives of the likelihood with respect to the parameters.
To avoid parameters becoming negative, we use log-transformed parameters $w = \log(x)$ for the parameter inference via MCMC.
We place a standard normal prior on the log-transformed parameters $w$.
For this dataset we choose MALA, based on the successful application and extensive study of this MCMC algorithm in previous work \citep{Riabiz2020, chen2019stein, Calderhead2009, Oates2016thermo}.
For each dataset we run five chains, where the initial conditions for each chain are sampled from the prior.
Each chain is run with a step size of $h=0.0033$ for $500000$ steps.
We use thinning with a step size of $20$.
This results in datasets of $n=125000$.
We did not use any warm-up on this dataset to keep the choice consistent for all dataset sizes. 

\begin{figure*}
	\includegraphics{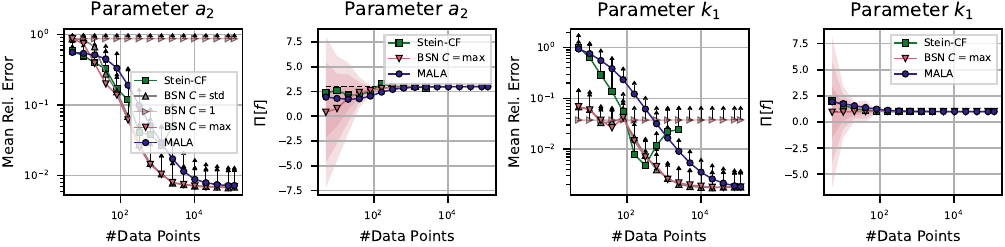}
	\caption{
  \emph{Posterior expectations for the parameters of a Goodwin ODE}.
  Mean relative integration error (\emph{left} and \emph{centre-right}), and uncertainty estimates (\emph{centre-left} and \emph{right}) (based on 5 repetitions) as a function of $n$. 
  }
	\label{fig:goodwin_oscillator2}
\end{figure*}

Figure~\ref{fig:goodwin_oscillator2} shows the results for the remaining two parameters not shown in the main text.

\subsection{Wind Farm Modelling}
\label{sec:exp_details_wind_farm}
For the wind farm model in our experiments, we assume we have a large-scale wind farm with equally spaced turbines on a two-dimensional grid and an ambient turbulence intensity. 
For each turbine, we use a wake deficit model by \citet{Niayifar2016}.
We put the following distributions on parameters for the wind farm simulation
\begin{itemize}
  \item \textbf{Turbine resistance coefficient:} Gaussian distribution with mean $\mu=1.33$ and variance $\sigma^2=0.1$.
  \item \textbf{Coefficient describing the wake expansion:} Gaussian distribution left-truncated at $0$ with mean $\mu=0.38$ and variance $\sigma = 0.001$.
  \item \textbf{Second coefficient describing the wake expansion:} Gaussian distribution left-truncated at $0$ with mean $\mu=4e-3$ and variance $\sigma^2=1e-8$.
  \item \textbf{Turbulence intensity:} Gaussian distribution left-truncated at $0$ with mean $\mu=0.1$ and variance $\sigma^2 = 0.003$
  \item \textbf{Wind direction:} Mixture of Gaussian distributions truncated so as to have support on $[0, 45]$ with means $\mu_1=0, \mu_2=22.5, \mu_3=33.75$ and variances $\sigma_1^2=50, \sigma_2^2=40, \sigma_3^2=8$.
  \item \textbf{Hub heights:} Gaussian distribution left-truncated at $0$ with mean $\mu=100$ and variance $\sigma^2=0.5$.
  \item \textbf{Hub diameter:} Gaussian distribution left-truncated at $0$ with mean $\mu=100$ and variance $\sigma^2=0.1$.
\end{itemize}

These distributions were chosen to have scales which might realistically represent uncertainty for their input, but if applying our method in practice these would have to be elicited from wind-farm experts. Note that the BSNs could be applied to much more complex distributions so-long as the density of $\Pi$ can be evaluated pointwise up to some normalization constant.

Our code is based on the code estimating the local turbine thrust coefficient \citet{Kirby2022} using a low-order wake model provided here: \url{https://github.com/AndrewKirby2/ctstar_statistical_model}. This code is based on the \texttt{PyWake} package \citep{pywake}.

\section{Bayesian Quadrature}
We now provide a short introduction to BQ and the derivation of the kernel mean embedding for truncated Gaussians.
\subsection{Introduction to Bayesian quadrature}
\label{sec:bq_details}

Recall that we are interest in approximating the integral $\Pi[f] = \int_{\mathcal{X}} f(x) \pi(x) dx$. BQ works by placing a $\mathcal{GP}(m, k)$ on $f$, i.e. a GP with mean function $m:\mathcal{X} \rightarrow \mathbb{R}$ and covariance functions (or kernel) $k : \mathcal{X}  \times \mathcal{X} \rightarrow \mathbb{R}$.
Then, given observation $\{x_i,f(x_i)\}_{i=1}^n$, we can compute the posterior mean and variance on the value of $\Pi[f]$ as
\begin{talign*}
	\mathbb{E}\left[ \Pi[f] \right] &= \int_{\mathcal{X}} \left(m(x) + k(x, x_{1:n}) k(x_{1:n},x_{1:n})^{-1} (f(x_{1:n})-m(x_{1:n}))\right) \pi(x) dx\\
	&= \Pi[m] + \Pi[k(\cdot, x_{1:n})] k(x_{1:n},x_{1:n})^{-1} (f(x_{1:n})-m(x_{1:n})),\\
	\mathbb{V}\left[ \Pi[f] \right] &=
	\int_{\mathcal{X}} \int_{\mathcal{X}} \left(k(x, x') - k(x, x_{1:n}) k(x_{1:n},x_{1:n})^{-1}  k(x_{1:n}, x')\right) \pi(x) \pi(x') dx dx'\\
	&= \Pi\bar{\Pi}[k] + \Pi[k(\cdot, x_{1:n})] k(x_{1:n},x_{1:n})^{-1} \Pi[k(x_{1:n}, \cdot)],
\end{talign*}
where $f(x_{1:n}) \in \mathbb{R}^n$ with $[f(x_{1:n})]_{i} = f(x_i)$, $m(x_{1:n}) \in \mathbb{R}^n$ with $[m(x_{1:n})]_{i} = m(x_i)$,  $k(x_{1:n},x)^\top = k(x,x_{1:n})\in \mathbb{R}^n$ with $[k(x,x_{1:n})]_i = k(x,x_{i})$,  $k(x_{1:n}, x_{1:n}) \in \mathbb{R}^{n \times n}$ with $[k(x_{1:n}, x_{1:n})]_{ij} = k(x_i,x_j)$ for all $i,j$ in $\{1,\ldots,n\}$. Finally, $\Pi\bar{\Pi}[k] = \int_{\mathcal{X}} \int_{\mathcal{X}} k(x, x')  \pi(x) \pi(x') dx dx'$.

Clearly, the expressions above can only be used if $\Pi[k(\cdot, x)]$, called the kernel mean embedding, and $\Pi\bar{\Pi}[k]$, called the initial error, are known in closed-form. This is only possible for some combinations of distribution $\pi$ and covariance function $k$. For example, if $\mathcal{X}=\mathbb{R}^d$, $\pi$ is a Gaussian and $k$ is the RBF-kernel, then the expressions above can be computed analytically. A more challenging case is that of truncated Gaussian distributions. In the next section, we show that the kernel mean embedding can be derived in that case.

\subsection{Kernel Mean Embedding for Truncated Gaussians}
\label{sec:bq_truncated_gaussians}
For truncated Gaussian distributions and the RBF kernel, we can compute the posterior mean but not the posterior variance.
Here we consider the 1-dimensional case with $\mathcal{X} = [a, b]$ which can be extended to the $d$-dimensinal case for isotropic Gaussians. 
We provide the expression for the kernel mean embedding: $\Pi[k(\cdot, x)] = \int_{\mathcal{X}} k(x', x) \pi(x')dx'$. We consider the case when $\pi$ is a truncated Gaussian and introduce the following notation:
\begin{talign*}
	\pi(x) = \frac{\phi(x, \mu, \sigma)}{\Phi(\frac{b-\mu}{\sigma}) - \Phi(\frac{a -\mu}{\sigma})}
\end{talign*}
where $\phi(x) = (\sqrt{2 \pi} \sigma)^{-1} \exp (-(x-\mu)^2/2\sigma^2)$ and  $\Phi(x) = \frac{1}{2}(1 + \text{erf}(x/\sqrt{2}))$. 

We use $Z$ to denote the normalization constant
\begin{talign*}
	Z(a,b,\mu, \sigma) = \Phi\left(\frac{b-\mu}{\sigma}\right) - \Phi\left(\frac{a -\mu}{\sigma}\right)
\end{talign*}
We rewrite the RBF kernel using the above identities $k(x, x') = \exp\left(-(x-x')^2/2l^2\right) = l \sqrt{2 \pi} \phi(x, x', l)$. We can now express the kernel mean embedding as:
\begin{talign*}
	\Pi[k(\cdot, x)] = \int_a^b  l \sqrt{2 \pi} \phi(x, x', l) \frac{\phi(x', \mu, \sigma)}{Z(a, b, \mu, \sigma)} dx' = C l \sqrt{2 \pi}\int_a^b \frac{\phi(x', \tilde{\mu}, \tilde{\sigma})}{Z(a, b, \mu, \sigma)} dx' = l \sqrt{2 \pi} C \frac{Z(a, b, \tilde{\mu}, \tilde{\sigma})}{Z(a, b, \mu, \sigma)},
\end{talign*}
where For truncated Gaussian distributions and the RBF kernel, we can compute the posterior mean but not the posterior variance.
Here we consider the 1-dimensional case with $\mathcal{X} = [a, b]$ which can be extended to the $d$-dimensinal case for isotropic Gaussians. 
We provide the expression for the kernel mean embedding: $\Pi[k(\cdot, x)] = \int_{\mathcal{X}} k(x', x) \pi(x')dx'$. We consider the case when $\pi$ is a truncated Gaussian and introduce the following notation:
\begin{talign*}
	\pi(x) = \frac{\phi(x, \mu, \sigma)}{\Phi(\frac{b-\mu}{\sigma}) - \Phi(\frac{a -\mu}{\sigma})}
\end{talign*}
where $\phi(x) = (\sqrt{2 \pi} \sigma)^{-1} \exp (-(x-\mu)^2/2\sigma^2)$ and  $\Phi(x) = \frac{1}{2}(1 + \text{erf}(x/\sqrt{2}))$. 

We use $Z$ to denote the normalization constant
\begin{talign*}
	Z(a,b,\mu, \sigma) = \Phi\left(\frac{b-\mu}{\sigma}\right) - \Phi\left(\frac{a -\mu}{\sigma}\right)
\end{talign*}
We rewrite the RBF kernel using the above identities $k(x, x') = \exp\left(-(x-x')^2/2l^2\right) = l \sqrt{2 \pi} \phi(x, x', l)$. We can now express the kernel mean embedding as:
\begin{talign*}
	\Pi[k(\cdot, x)] = \int_a^b  l \sqrt{2 \pi} \phi(x, x', l) \frac{\phi(x', \mu, \sigma)}{Z(a, b, \mu, \sigma)} dx' = C l \sqrt{2 \pi}\int_a^b \frac{\phi(x', \tilde{\mu}, \tilde{\sigma})}{Z(a, b, \mu, \sigma)} dx' = l \sqrt{2 \pi} C \frac{Z(a, b, \tilde{\mu}, \tilde{\sigma})}{Z(a, b, \mu, \sigma)},
\end{talign*}
where 
\begin{talign*}
	\tilde{\mu} = \frac{\mu l^2 + x \sigma^2}{\sigma^2 + l^2}, \qquad 
	\tilde{\sigma} = \sqrt{\frac{\sigma^2 l^2}{\sigma^2 + l^2}}, \qquad
	C = \frac{1}{\sqrt{2 \pi (\sigma^2 + l^2)}}\exp\left(\frac{(\mu - x)^2}{2 (\sigma^2 + l^2)}\right).
\end{talign*}
\subsection{Kernel Mean Embedding for Matern 1/2 Kernel}
\label{sec:matern}
For Gaussian distributions and the Matern 1/2 kernel, we can compute the posterior mean but only in $d=1$.
We provide the expression for the kernel mean embedding: $\Pi[k(\cdot, x)] = \int_{\mathbb{R}} k(x', x) \pi(x')dx'$, where $\pi(x) = \mathcal{N}(0, 1)$ is a standard normal and $k(x', x) = \exp(|x - x'|/l)$ is the Matern 1/2 kernel.
\begin{talign*}
	\Pi[k(\cdot, x)]=\frac{1}{2} \exp \left(\frac{2 x l + 1}{2l^2}\right) \mathrm{erfc} \left(\frac{x + \frac{1}{l}}{\sqrt{2}}\right) + \frac{1}{2} \exp \left(\frac{1 - 2xl}{2l^2}\right) \left(\mathrm{erf}\left(\frac{x - \frac{1}{l}}{\sqrt{2}}\right) + 1 \right)
\end{talign*}

\section{ADDITIONAL BACKGROUND: LAPLACE APPROXIMATION}
\label{sec:ggn_approx}
The Laplace approximation constructs a second-order Taylor approximation around the maximum of the posterior, i.e., the mode of the posterior, which amounts to a Gaussian approximate of the posterior around the MAP (maximum a-posteriori) estimate.
Here we provide a detailed introduction. 

We want to compute a posterior for the parameters of our model
\begin{talign}
  \begin{split}
  p(\theta| \mathcal{D}) &= \frac{p(\mathcal{D}| \theta)p(\theta)}{Z},\quad \text{where}\\
  Z &= \int p(\mathcal{D}| \theta)p(\theta) d\theta.
  \end{split}
  \label{eq:posterior}
\end{talign}
Here, the integral for the normalization constant $Z$ is usually not tractable, and we will have to resort to some approximation to compute it.
We provide the expressions for negative log prior
\begin{talign}
  - \log p(\theta) = \frac{1}{2 \sigma_0^2} \|\theta\|_2^2 - \frac{p+1}{2}\log \pi \sigma_0^2
  \label{eq:prior}
\end{talign}
and the negative log likelihood
\begin{talign}
  - \log p(\mathcal{D}|\theta) = \frac{1}{2 \sigma^2} \sum_{i=1}^n \|f(x_i) - g_\theta(x_i)\|_2^2 - \frac{n}{2}\log \pi \sigma^2,
  \label{eq:likelihood}
\end{talign}
where $\sigma$ is the dataset noise.
By comparing \eqref{eq:likelihood} and \eqref{eq:prior} to the mean square loss with weight decay
\begin{talign*}
    \begin{split}
        l_\text{tot}(\theta) & = l(\theta) + \lambda \|\theta\|_2^2 \\
        l(\theta) & = \frac{1}{n} \sum_{i=1}^n \|f(x_i) - g_\theta(x_i)\|_2^2
    \end{split}
\end{talign*}
we note $l \propto - \log p(\mathcal{D}| \theta)$ and $\lambda \|\theta\|_2^2 \propto - \log p(\theta)$.
Hence, the minimum of the loss correspond the maximum of the posterior, i.e. $\theta_{\text{MAP}} = \text{argmin}_{\theta} l_{\text{tot}}(\theta) = \text{argmin}_{\theta} -\log p(\mathcal{D}| \theta) - \log p(\theta) = \text{argmin}_{\theta} -\log p(\theta| \mathcal{D})$.  
We denote $L(\theta)  = \log p(\mathcal{D}| \theta) + \log p(\theta)$, and rewrite Equation \eqref{eq:posterior} 
\begin{talign*}
  p(\theta| \mathcal{D}) = \frac{e^{L(\theta)}}{Z}.  
\end{talign*}
To find a suitable approximation for the posterior we, we use as Taylor series expansion of $L$ around $\theta_{\text{MAP}}$
\begin{talign*}
  L(\theta) \approx L(\theta_\text{MAP}) + (\theta - \theta_\text{MAP})^\top \nabla_{\theta}L(\theta_{\text{MAP}})  + \frac{1}{2}(\theta - \theta_\text{MAP})^\top \nabla_{\theta}^2 L(\theta_{\text{MAP}}) (\theta - \theta_\text{MAP})
\end{talign*}
The second term is equal to zero by definition of $\theta_\text{MAP}$.
Hence, we arrive at a Gaussian approximation of the posterior $q(\theta)$ for the form
\begin{talign*}
 q_{\text{Laplace}}(\theta) = \mathcal{N}\left(\theta \mid \theta_\text{MAP}, \Sigma \right),
\end{talign*}
where $\Sigma$ is proportional to the inverse Hessian of the loss $l_{\text{tot}}$ evaluated at $\theta_\text{MAP}$: 
\begin{talign*}
 \Sigma^{-1} &=\nabla_{\theta}^2 L|_{\theta=\theta_{\text{MAP}}} 
 =  \left(- \nabla_\theta^2 \log p(\mathcal{D}|\theta) - \nabla_\theta^2 \log p(\theta)\right)|_{\theta=\theta_{\text{MAP}}}\\
 &= H  +  \sigma_0^{-2} I_{p+1}.
\end{talign*}
Since computing $H$ is computationally expensive, we use the GGN Approximation to compute it.

\subsection{GGN Approximation}
For the Laplace approximation we are interested in computing the Hessian of the log likelihood $-\log p(\mathcal{D}| \theta) \propto \frac{1}{2 \sigma^2} \sum_{i=1}^n \|f(x_i) - g_\theta(x_i)\|_2^2$.
The GGN approximation is a positive-semi-definite approximation of the full Hessian $H$, i.e.,
\begin{talign*}
  H &=  \nabla_{\theta}^2 \left(-\log p(\mathcal{D}| \theta)\right)\\
    &=  \nabla_{\theta}^2 \frac{1}{2 \sigma^2} \sum_{i=1}^n \|f(x_i) - g_\theta(x_i)\|_2^2 \\
    &= \frac{1}{2 \sigma^2} \sum_{i=1}^n \left[ \left(\nabla_{\theta} g_{\theta}(x_i)|_{\theta=\theta_{\text{MAP}}}\right)^\top 
    2 
    \nabla_{\theta} g_{\theta}(x_i)|_{\theta=\theta_{\text{MAP}}}
  + \nabla_{\theta}^2 g_{\theta}(x_i)|_{\theta=\theta_{\text{MAP}}} 
  \left(\nabla_g \|f(x_i) - g\|_2^2|_{g=g_{\theta_{\text{MAP}}}(x_i)}\right)^2 \right]\\
  &= H_{\text{GGN}} + R,
\end{talign*}
where $\sigma$ is the dataset noise as in Eq. \eqref{eq:likelihood}.
The GGN approximation is given by
\begin{talign*}
  H_{\text{GGN}} = \frac{1}{\sigma^2}\sum_{i=1}^n J(x_i)^\top J(x_i)
\end{talign*}
where $J(x_i) = \nabla_{\theta} g_{\theta}(x_i)|_{\theta=\theta_{\text{MAP}}}$.

\end{appendices}

\end{document}

%% file: tikz/sketch_methods.tikz
\tikzset{
	neuron/.style={ 
		rectangle,draw, line width=0.5mm, rounded corners, 
	},
}

\begin{tikzpicture}
\node[inner sep=0pt] (f_pdf) {\includegraphics[]{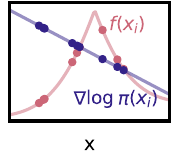}};
\node[right=0.5em of f_pdf](f_node2){};
\node[above=1em of f_node2](f_node){};
\node[right=3em of f_node](regression_node){};
\node[below=.5em of regression_node](regression_node2){};
\node[right=2em of regression_node2](regression_node4){};
\node[right=6em of regression_node2](regression_node3){};
\node [neuron, below =5em of regression_node, align=left](theta_map) {$\theta_{\text{MAP}}$};
\node [neuron, right=0.5em of f_node, align=left](regression) {Neural network regression  for $g_\theta$
\\$\{x_i, f(x_i), \nabla \log \pi(x_i)\}$};
\node[inner sep=0pt, below=1em of regression_node3] (g_fit) {\includegraphics[]{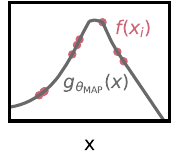}};
\node [neuron, below=10em of regression_node4, align=left](laplace) {Laplace approximation};
\node[left=0.5em of laplace](g_node){};
\node[above=0.5em of g_node](g_node2){};
\node [inner sep=0pt, left= 1em of g_node2](theta_png){\includegraphics[]{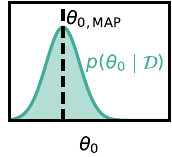}};

\draw[line width=0.5mm,-latex] let \p1=(f_pdf.east),\p2=(regression.west) in (\x1,\y2)--(\x2,\y2);
\draw[line width=0.5mm,-latex] let \p1=(regression.south),\p2=(theta_map.north) in (\x2,\y1)--(\x2,\y2);
\draw[line width=0.5mm,-latex] let \p1=(regression.south),\p2=(g_fit.north) in (\x2,\y1)--(\x2,\y2);
\draw[line width=0.5mm,-latex] let \p1=(theta_map.south),\p2=(laplace.north) in (\x1,\y1)--(\x1,\y2);
\draw[line width=0.5mm,-latex] let \p1=(laplace.west),\p2=(theta_png.east) in (\x1,\y1)--(\x2,\y1);
\end{tikzpicture}